\documentclass{article}

\usepackage[main, final]{neurips_2026}

\usepackage[utf8]{inputenc} 
\usepackage[T1]{fontenc}    
\usepackage{hyperref}       
\usepackage{url}            
\usepackage{booktabs}       
\usepackage{amsfonts}       
\usepackage{nicefrac}       
\usepackage{microtype}      
\usepackage{xcolor}         
\usepackage{amsmath}

\usepackage{tcolorbox}
\usepackage{tcolorbox}
\tcbuselibrary{skins,breakable}
\usepackage{booktabs}    
\usepackage{graphicx}    
\usepackage{amssymb}     
\usepackage{bbding}      
\usepackage{algorithm}
\usepackage{algpseudocode}
\usepackage{amsmath}
\usepackage{color}
\usepackage{multirow}
\usepackage{booktabs}
\usepackage{multirow}
\usepackage{xcolor}
\usepackage{colortbl}
\usepackage{wrapfig}
\usepackage{makecell}
\usepackage{enumitem}

\usepackage{subcaption}

\newif\ifshowcomments
\showcommentstrue

\ifshowcomments

\title{MedZERO: Self-Evolving Agents for \\ Open-Ended Medical Reasoning Through \\Controlled Knowledge Accumulation}

\author{%
  \textbf{Xilin Dang}$^{1}$ \quad
  \textbf{Weilin Ruan}$^{1}$ \quad
  \textbf{Xue Yang}$^{2}$ \quad
  \textbf{Jinghao Wang}$^{1}$ \\
  \textbf{Xiaowei Hu}$^{3}$\thanks{Corresponding authors:
  Xiaowei Hu (\texttt{huxiaowei@scut.edu.cn}) and
  Jinpeng Li (\texttt{jinpeng.li@cair-cas.org.hk}).} \quad
  \textbf{Jinpeng Li}$^{4}$\footnotemark[1] \quad
  \textbf{Pheng-Ann Heng}$^{1}$ \\[0.5em]
  \small $^{1}$Department of Computer Science and Engineering,
  \small The Chinese University of Hong Kong\\
  \small $^{2}$Department of Hematology, Zhongshan Hospital, Fudan University\\
  \small $^{3}$School of Future Technology, South China University of Technology\\
  \small $^{4}$Center for Artificial Intelligence and Robotics,\\
  \small Hong Kong Institute of Science and Innovation, Chinese Academy of Sciences
}

\begin{document}

\maketitle

\begin{abstract}
Large language models (LLMs) have shown promise in medical question answering and clinical reasoning, yet their improvement remains constrained by static parametric knowledge and costly expert supervision.
Self-evolving agents offer a promising alternative by enabling models to improve through iterative task generation and problem-solving.
However, most existing self-evolving methods are designed for easily verifiable domains such as mathematics and coding, where solutions can be checked by exact answers or executable programs.
Medical reasoning is fundamentally different: it is open-ended, knowledge-intensive, and often only partially verifiable.
We present \textbf{MedZERO}, a self-evolving framework for open-ended medical reasoning.
MedZERO couples an \textit{Examiner} that generates frontier medical question-option pairs with a \textit{Reasoner} that solves them through evidence-grounded multi-turn reasoning with external knowledge tools.
To support reliable, continual improvement, MedZERO adopts controlled knowledge accumulation, which maintains temporary exploratory knowledge and curated persistent knowledge in reasoning.
We evaluate MedZERO on five public medical reasoning benchmarks using 4B- and 8B-scale base models under open-ended evaluation.
At both model scales, MedZERO achieves the highest average
accuracy among the evaluated methods.
Code is available at \url{https://github.com/IcecreamArtist/MedZERO}.
\end{abstract}

\section{Introduction}

\begin{figure}[ht]
\centering
\includegraphics[width=1\textwidth]{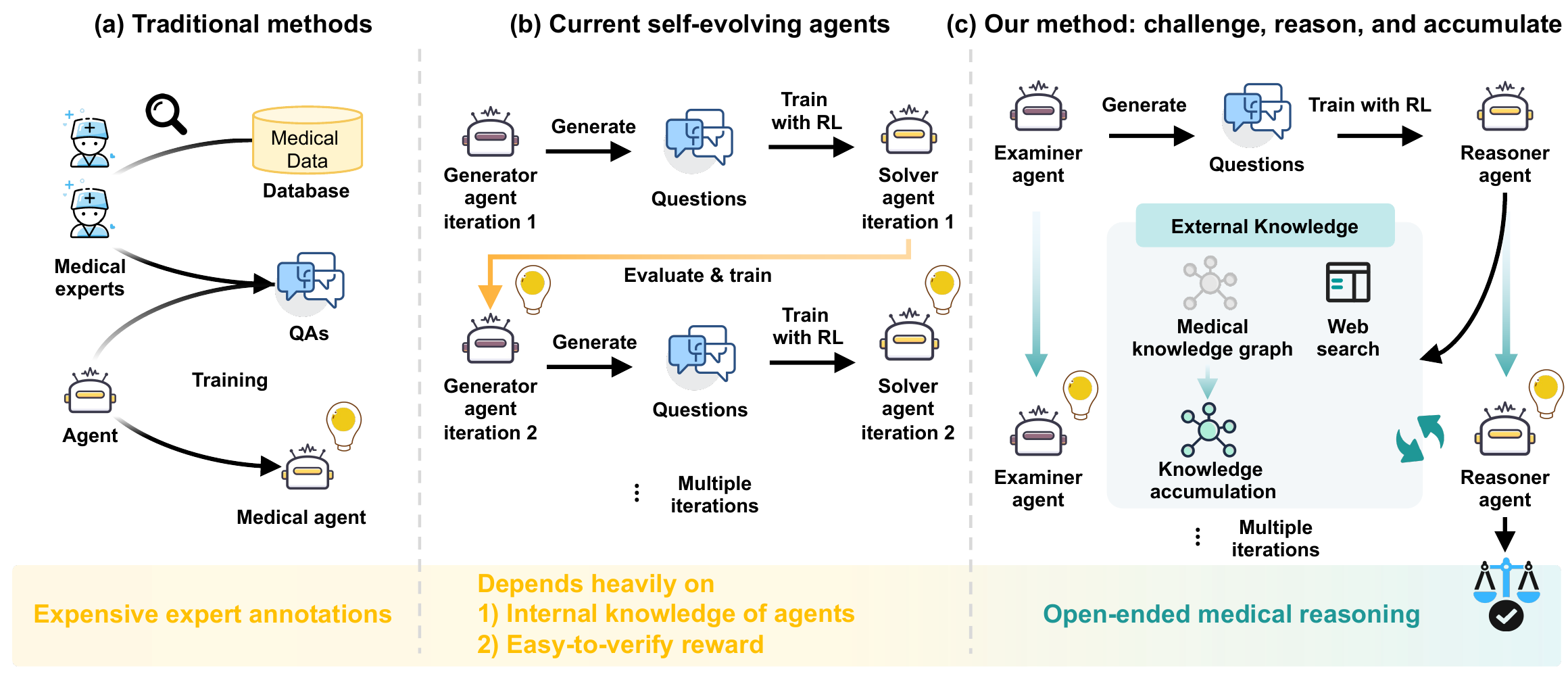}
\caption{(a) Conventional medical reasoning pipelines rely on expensive expert supervision and static knowledge. (b) Prior self-evolving methods have shown their strongest results in easily verifiable domains such as mathematics and coding. (c) MedZERO extends self-evolution to open-ended medical reasoning through Examiner--Reasoner co-evolution, evidence-grounded tool use, and controlled knowledge accumulation.}
\label{fig:teaser}
\end{figure}

Large language models (LLMs) have shown strong potential in medical question answering, evidence synthesis, and complex clinical reasoning~\cite{singhal2025toward,jin2021disease,pal2022medmcqa}. 
However, their medical reasoning ability remains limited by two practical bottlenecks: static parametric knowledge and expensive expert supervision. 
In medicine, obtaining high-quality supervision is particularly costly because it often requires domain experts, careful quality control, and substantial time for adjudicating difficult cases~\cite{johri2025evaluation}. 
These constraints make it difficult to scale supervised fine-tuning and reinforcement learning pipelines for advanced medical reasoning~\cite{ouyang2022training,guo2025deepseek,dang2026advancing,dang2026hivlr}.

Recent self-evolving frameworks offer a promising alternative by allowing models to generate tasks, explore solution strategies, and improve through iterative self-improvement~\cite{madaan2023self,shinn2023reflexion,zelikman2022star,yuan2024self}. 
Yet their strongest results have so far been concentrated in domains such as mathematics and coding, where candidate solutions can be checked by exact answers or executable programs~\cite{zhao2025absolutezeroreinforcedselfplay,liu2025spiral,wang2025socratic,huang2025rzeroselfevolvingreasoningllm}. 
Medical reasoning is fundamentally different. First, medical questions often admit free-form explanations, differential diagnoses, or evidence-dependent conclusions, so semantically correct answers may differ substantially in wording. Second, unlike math or coding, many medical reasoning steps do not have an exact executable verifier; automatic correctness signals are therefore weak, partial, or dependent on imperfect judges. Third, medical progress often requires acquiring, organizing, and reusing external evidence, such as biomedical facts, clinical guidelines, or recently retrieved information, rather than only recombining knowledge already stored in model parameters~\cite{tao2024survey}.
As a result, extending self-evolution from easily verifiable domains to medicine is not a straightforward transfer, but a distinct problem setting.

These task properties create three algorithmic challenges for self-evolution.
First, self-generated questions often remain too close to the model's existing comfort zone, making it hard to sustain a useful frontier curriculum over time~\cite{shi2025efficient,bae2025online}. 
This problem is especially severe in medicine, where new progress often requires bringing in external knowledge rather than simply resampling familiar reasoning patterns. 
Second, medical reasoning rarely admits exact automatic verification, so self-consistency alone is not a reliable training signal for open-ended answers~\cite{zheng2023judging,dang2026knowguard}. 
Third, useful knowledge discovered during self-evolution is difficult to preserve safely: naively feeding all self-generated facts, intermediate hypotheses, or retrieved snippets back into future reasoning can accumulate unsupported artifacts and reinforce errors. 
Although retrieval-augmented and search-based medical agents improve evidence access and answer grounding~\cite{jin2025search,chandak2023building,lewis2020retrieval}, they do not by themselves solve these challenges of curriculum construction, partial verifiability, and controlled knowledge reuse.

To address this gap, we present \textbf{MedZERO}, a self-evolving framework for open-ended medical reasoning. 
MedZERO consists of two co-evolving agents. An \textit{Examiner} generates medical question-option pairs near the current capability frontier of the learner, providing a weak but usable frontier signal under partial verifiability. A \textit{Reasoner} then solves these questions through multi-turn reasoning with external knowledge tools, including structured retrieval and optional web search. 
To stabilize learning in this open-ended setting, MedZERO contains three ingredients: \emph{frontier curriculum generation}, \emph{joint majority-consistency and evidence-grounded supervision}, and \emph{controlled knowledge accumulation}. 
In particular, MedZERO distinguishes transient exploratory knowledge produced during problem solving from curated persistent knowledge that is admitted into the visible retrieval substrate. This separation allows broad exploration while reducing the risk that unsupported self-generated artifacts directly contaminate future reasoning and training.

We evaluate MedZERO on five public benchmarks for open-ended, knowledge-intensive medical reasoning, reporting MedBullets under two option settings and averaging over all six reported metrics, using both 4B- and 8B-scale base models. 
At both model scales, MedZERO achieves the highest average
accuracy among the evaluated methods, with gains of up to
13.7 percentage points over the next-best self-evolving baseline.
These results show that self-evolution can be extended beyond easily verifiable domains when curriculum construction, supervision, and knowledge accumulation are explicitly designed for partial verifiability and continual knowledge acquisition.

Our contributions are summarized as follows:
\begin{itemize}[leftmargin=*]

\item We propose \textbf{MedZERO}, a dual-agent framework in which an \textit{Examiner} generates frontier medical question-option pairs and a \textit{Reasoner} improves through evidence-grounded multi-turn reasoning with external knowledge tools.

\vspace{-0.5mm}
\item We introduce \textit{controlled knowledge accumulation}, which separates transient exploratory knowledge from curated persistent knowledge visible to the Reasoner, enabling continual improvement while mitigating contamination from unsupported self-generated artifacts.

\vspace{-0.5mm}
\item We validate MedZERO on five medical reasoning benchmarks across 4B- and 8B-scale backbones under open-ended evaluation, obtaining up to 22.7 average-point gains over the base model and 18.7 points over prior self-evolving baselines.

\end{itemize}

\section{Related Work}

\subsection{Self-Evolution from Zero Data}

Recent work has explored self-evolving or self-improving language model frameworks that reduce reliance on curated supervision by generating training tasks and trajectories autonomously~\cite{zelikman2022star,madaan2023self,shinn2023reflexion}. These approaches have shown the strongest results in verifiable domains such as coding and mathematics~\cite{guo2025deepseek,shao2024deepseekmath}, where learning can be guided by execution feedback, deterministic answers, confidence, or consistency-based proxy rewards~\cite{yuan2024self}. 
More recent work extends self-evolution to fully autonomous systems that generate and solve their own problems from scratch~\cite{xia2025agent0,zhao2025absolutezeroreinforcedselfplay,huang2025rzeroselfevolvingreasoningllm,wang2025socratic,lu2025search}. However, such signals become much less reliable in open-ended settings for reinforcement learning techniques such as Group Relative Policy Optimization (GRPO)~\cite{shao2024deepseekmath}, where correctness is ambiguous and self-generated supervision can reinforce existing errors~\cite{tao2024survey,tu2025position,yue2025does}.
MedZERO targets this gap in the medical domain. Unlike prior self-evolving methods designed for easily verifiable tasks, MedZERO addresses open-ended medical reasoning under partial verifiability.

\subsection{Interactive Medical Reasoning}

Most prior work on medical language models focuses on static question answering benchmarks~\cite{singhal2025toward,jin2019pubmedqa,pal2022medmcqa,chen2025benchmarking,zuo2025medxpertqa}, where the full clinical context is given in a single prompt. More recent work studies interactive medical reasoning, in which models must decide what information to acquire or what questions to ask before answering. MediQ~\cite{li2024mediq} and KnowGuard~\cite{dang2026knowguard} evaluate question-asking LLMs for interactive clinical reasoning. AgentClinic~\cite{schmidgall2024agentclinic} tests agents in simulated clinical environments. CRAFT-MD~\cite{johri2025evaluation} assesses conversational diagnosis through doctor-patient interaction. Multi-agent medical architectures such as MedAgents~\cite{tang2024medagents} and MDAgents~\cite{kim2024mdagents} demonstrate that collaboration structures can improve medical reasoning without additional training data. Agent Hospital~\cite{li2024agent} explores an alternative where doctor agents evolve through simulated patient interactions without labeled data. While these settings better reflect clinical reasoning, they still typically rely on curated datasets or fixed environments rather than autonomous self-improvement.
MedZERO complements this line of work by introducing a self-evolving framework for interactive, knowledge-intensive medical reasoning.
\begin{figure}[t]
\centering
\includegraphics[width=1\textwidth]{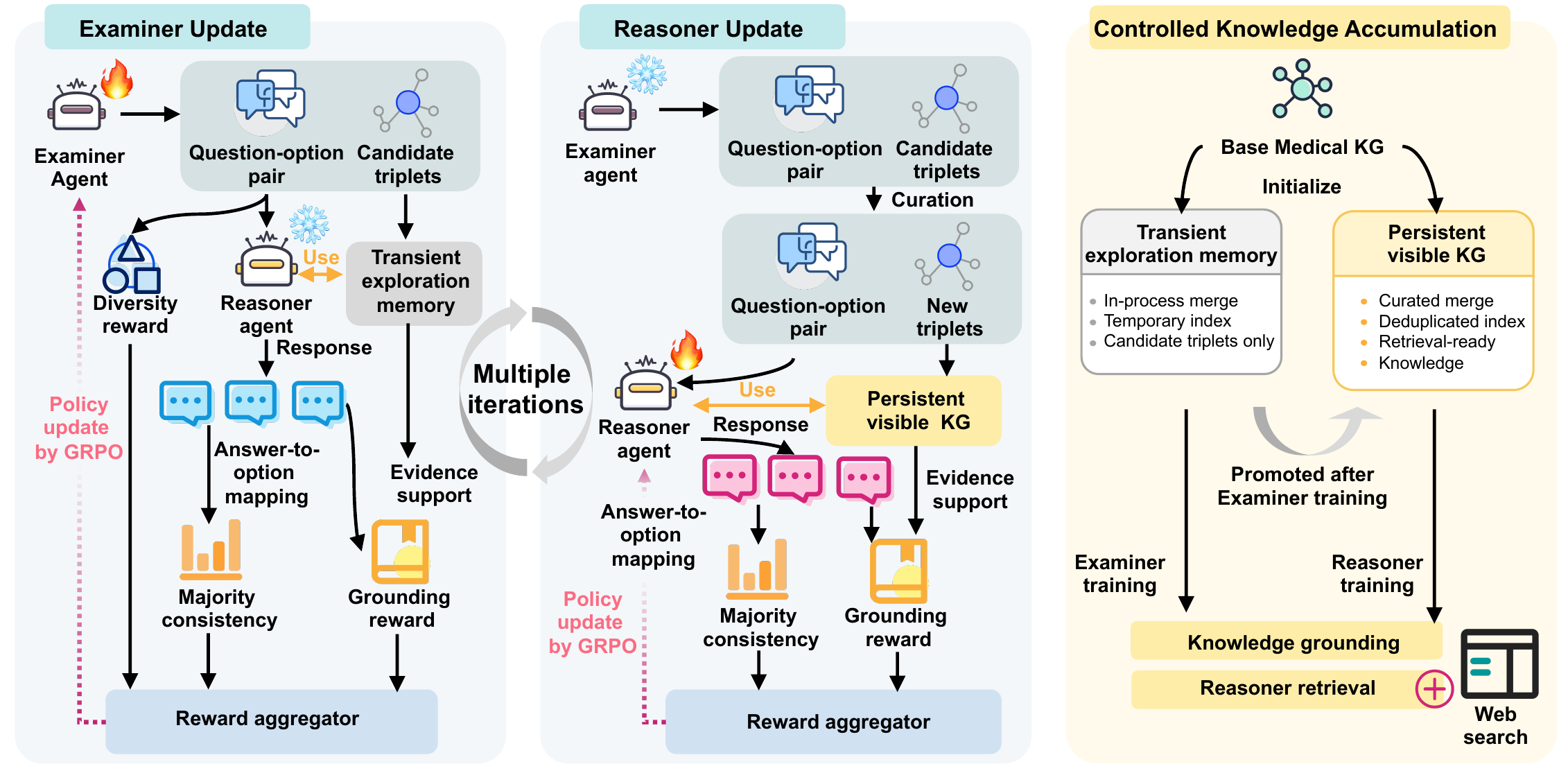}
\caption{
Overview of MedZERO. MedZERO is a self-evolving framework for open-ended medical reasoning that couples frontier curriculum generation, evidence-grounded reasoning, and controlled knowledge accumulation. An \textbf{Examiner} generates medical question-option pairs and candidate knowledge triplets near the current capability frontier of the \textbf{Reasoner}. The Reasoner solves these questions with structured retrieval and optional web search. MedZERO separates exploratory triplets from curated persistent knowledge, enabling knowledge accumulation without directly reinforcing unverified self-generated artifacts.
}
\label{fig:overview}
\end{figure}

\section{Methodology}

\subsection{Framework Overview}
\label{sec:overview}

MedZERO is a dual-agent self-evolving framework for open-ended medical reasoning. It consists of an \textbf{Examiner}, which generates training questions with local candidate option sets and candidate medical knowledge triplets, and a \textbf{Reasoner}, which answers them using multi-turn reasoning with external knowledge tools. MedZERO trains the two agents through alternating self-evolution rounds, where each round consists of an Examiner update and a Reasoner update.

At round $t$, MedZERO first updates the Examiner while freezing the Reasoner. The frozen Reasoner evaluates Examiner-generated question-option pairs to optimize the Examiner. During this stage, the Examiner also generates candidate triplets, which may be used as temporary exploration-time knowledge but are not directly promoted to the persistent retrieval knowledge base.
MedZERO then freezes the updated Examiner. The Examiner regenerates candidate question-option-triplet samples, and 
samples with at least one non-abstained answer mapping and an option-agreement score within the frontier band are retained.
Their question-option pairs form the Reasoner training curriculum, while their associated triplets are deduplicated and accumulated into the persistent generated knowledge graph.
Finally, MedZERO updates the Reasoner while freezing the Examiner. Repeating these stages over multiple rounds yields co-evolution: a stronger Reasoner shifts the frontier used to train the Examiner, and the improved Examiner produces harder and better-grounded data for subsequent Reasoner training.

\subsection{Frontier Curriculum Generation}
\label{sec:curriculum}

A useful self-evolving curriculum should concentrate on questions near the Reasoner's current capability frontier. In open-ended medical reasoning, however, difficulty cannot be estimated by exact answer matching because free-form answers may be semantically equivalent but lexically diverse. MedZERO addresses this by letting the Examiner generate a question together with a local option set, which partially canonicalizes the answer space and enables option-level agreement to serve as a weak frontier signal.

Let $\pi_X$ denote the Examiner policy. Given the current Reasoner, the Examiner generates a question-option pair and associated candidate triplets,
\begin{equation}
(q, O, T) \sim \pi_X(\cdot),
\end{equation}
where $O=\{o_1,\dots,o_M\}$ is a local option set and $T$ is a set of generated medical knowledge triplets. 

The Examiner receives the reward
\begin{equation}
r_{\text{exam}}(q, O)
=
\lambda_{\text{front}} r_{\text{front}}(q, O)
+
\lambda_{\text{div}} r_{\text{div}}(q, O)
+
\lambda_{\text{ground}} r_{\text{ground}}(q),
\end{equation}
which will be introduced next, and is optimized by GRPO
\begin{equation}
\max_{\pi_X}
\;
\mathbb{E}_{(q,O,T) \sim \pi_X}
\left[
r_{\text{exam}}(q, O)
\right].
\end{equation}

\paragraph{Option-level frontier reward.}
For each generated question $q$, the Reasoner samples $K$
independent free-form responses $\{y_k(q)\}_{k=1}^{K}$,
optionally using knowledge retrieval but without observing
the options. A mapper then maps each response
to the Examiner-generated option set:
\begin{equation}
\label{eq:reason_map}
m(y_k(q), O) \in \{o_1,\dots,o_M,\varnothing\},
\end{equation}
where $\varnothing$ denotes an invalid or unparseable
mapping, which is excluded from the agreement calculation.

Let $n_j(q)$ be the number of responses mapped to option $o_j$, and let
\begin{equation}
n_{\mathrm{valid}}(q)=\sum_{j=1}^{M} n_j(q)
\end{equation}
be the number of non-abstained mappings. The option-level agreement score is
\begin{equation}
\label{eq:agreement}
s_{\text{agr}}(q, O)
=
\begin{cases}
\displaystyle \max_{j \in \{1,\dots,M\}} \frac{n_j(q)}{n_{\mathrm{valid}}(q)}, & n_{\mathrm{valid}}(q) > 0,\\[8pt]
0, & n_{\mathrm{valid}}(q)=0.
\end{cases}
\end{equation}
We define the frontier reward as
\begin{equation}
r_{\text{front}}(q, O)
=
-
\left|
s_{\text{agr}}(q, O) - \gamma
\right|,
\end{equation}
where $\gamma \in (0,1)$ is the target agreement level. Very high agreement indicates that the question is likely too easy, whereas very low agreement often indicates that the question is either too difficult or poorly specified. The reward therefore favors intermediate option-level agreement near the Reasoner's capability frontier.

\paragraph{Diversity reward.}
To avoid repeatedly generating near-duplicate questions, following~\cite{huang2025rzeroselfevolvingreasoningllm,xia2025agent0}, $r_{\text{div}}$ rewards question-option pairs that are dissimilar to previously retained high-reward samples in the curriculum buffer. This term serves as a standard anti-collapse regularizer.

\paragraph{Grounding reward.}
Frontier difficulty alone does not ensure medically meaningful supervision. Let $\tau(q)$ denote the Reasoner's trajectory for question $q$, and let $\mathcal{E}(q)$ denote the Reasoner's retrieved evidence, such as structured facts and web references. We define
\begin{equation}
r_{\text{ground}}(q)
=
s_{\text{sup}}\!\left(\tau(q), \mathcal{E}(q)\right),
\end{equation}
where $s_{\text{sup}}$ measures how well the trajectory is supported by retrieved evidence.

\subsection{Ephemeral-to-Persistent Knowledge Accumulation}
\label{sec:kg}

A central risk in medical self-evolution is uncontrolled knowledge feedback: if every Examiner-generated triplet is immediately stored as persistent retrieval knowledge, noisy or reward-hacking artifacts may be repeatedly retrieved and reinforced in later Reasoner training. MedZERO therefore separates optimization-time ephemeral triplets from post-optimization persistent knowledge.

Let $\mathcal{G}_0$ denote the base medical knowledge graph, and let $\mathcal{G}^{\mathrm{pers}}_{t}$ denote the persistent generated KG after round $t$. During Examiner optimization round $t$, candidate triplets produced by the Examiner are stored in a temporary buffer $\mathcal{G}^{\mathrm{tmp}}_t$. To avoid using information generated by the current instance as its own evidence, MedZERO applies a causal visibility rule. When processing step or batch $i$, the Reasoner can retrieve from the base KG and previously committed temporary triplets, but not from triplets generated by the current step or batch:
\begin{equation}
\mathcal{G}^{\mathrm{cur}}_{t,i}
=
\mathcal{G}_0
\cup
\mathcal{G}^{\mathrm{pers}}_{t-1}
\cup
\bigcup_{j<i} \mathcal{G}^{\mathrm{tmp}}_{t,j}.
\end{equation}
After Examiner optimization finishes, $\mathcal{G}^{\mathrm{tmp}}_t$ is discarded and is not directly promoted to the persistent KG.

Persistent accumulation is performed only through a separate post-optimization generation-and-curation pass. The optimized Examiner from round $t$ regenerates candidate samples
\begin{equation}
x=(q,O,T),
\end{equation}
where $T$ denotes the associated generated triplets. The current Reasoner evaluates each generated question--option
pair using the option-level agreement score defined in
Eq.~\eqref{eq:agreement}:
$s(x)=s_{\mathrm{agr}}(q,O)$.
This score measures agreement among valid mapped responses,
rather than accuracy against a reference answer.
We retain samples with at least one valid mapped response
and an agreement score within the frontier band:
\begin{equation}
D_t =
\left\{
x=(q,O,T):
s_{\min}\leq s_{\mathrm{agr}}(q,O)\leq s_{\max},
\ n_{\mathrm{valid}}(q)>0
\right\}.
\end{equation}
In our experiments, $s_{\min}=0.3$ and $s_{\max}=0.8$.

Triplets associated with retained samples are deduplicated and appended to the persistent generated KG:
\begin{equation}
\mathcal{G}^{\mathrm{pers}}_t
=
\mathcal{G}^{\mathrm{pers}}_{t-1}
\cup
\mathrm{Dedup}
\left(
\bigcup_{x \in \mathcal{D}_t} T(x)
\right).
\end{equation}
The retrieval-visible KG for subsequent Reasoner training and evaluation is therefore
\begin{equation}
\mathcal{G}^{\mathrm{vis}}_t
=
\mathcal{G}_0
\cup
\mathcal{G}^{\mathrm{pers}}_t.
\end{equation}
Thus, Examiner optimization may use generated triplets as ephemeral exploration-time memory, but only regenerated triplets that pass post-optimization filtering become persistent retrieval-visible knowledge.

\subsection{Reasoner Training}
\label{sec:executor}

The Reasoner is trained on the retained curriculum $\mathcal{D}_t$ using a reward that combines option-level majority consistency with evidence grounding. The former provides a weak self-consistency signal in the examiner-induced answer space, while the latter constrains the trajectory to be supported by retrieved medical evidence.

Let $\pi_R$ denote the Reasoner policy. For each training instance $(q,O,T) \in \mathcal{D}_t$, the Reasoner samples $K$ trajectories, optionally using the retrieval-visible KG $\mathcal{G}^{\mathrm{vis}}_t$ and web search:
\begin{equation}
\tau_k \sim \pi_R(\cdot \mid q, \mathcal{G}^{\mathrm{vis}}_t, \text{web\_search}), 
\qquad
k=1,\dots,K.
\end{equation}
Each trajectory produces a free-form final answer $y_k$, which is mapped into the local option space, in the same way as in Eq.~\eqref{eq:reason_map}: $m(y_k(q), O) \in \{o_1,\dots,o_M,\varnothing\}$.

The majority option among valid mapped responses is
\begin{equation}
o^{\star}(q)
=
\arg\max_{o_j \in O}
\sum_{k=1}^{K}
\mathbf{1}\!\left[m(y_k(q), O)=o_j\right].
\end{equation}
If no valid mapped response exists, the majority-consistency reward is set to zero. Otherwise, for each trajectory $\tau_k$,
\begin{equation}
r_{\text{maj}}(\tau_k)
=
\mathbf{1}\!\left[m(y_k(q),O)=o^{\star}(q)\right].
\end{equation}

The Reasoner is jointly optimized by the majority-consistency reward and the grounding reward:
\begin{equation}
\max_{\pi_R}
\;
\mathbb{E}_{(q,O,T)\sim \mathcal{D}_t,\;\tau_k \sim \pi_R}
\left[
r_{\text{maj}}(\tau_k)
+
\beta r_{\text{ground}}(\tau_k)
\right].
\end{equation}

\subsection{Open-ended Evaluation}
\label{sec:eval}

Since MedZERO is trained to produce free-form medical answers, evaluation preserves the open-ended inference setting rather than directly decoding benchmark options.

For multiple-choice medical benchmarks, we use a two-stage protocol. Given context $c$ and question stem $x$ without answer options, the Reasoner first produces an open-ended response
\begin{equation}
y = \pi_R^{\text{open}}(c, x).
\end{equation}
A separate answer mapper maps $y$ to one option in the benchmark option set $\mathcal{O}$, which is the same as in Eq.~\eqref{eq:reason_map}:
\begin{equation}
\hat{\ell} = J(y, \mathcal{O}).
\end{equation}
Accuracy is computed against the gold label $\ell^\star$:
\begin{equation}
\mathrm{Acc}
=
\frac{1}{N}
\sum_{i=1}^{N}
\mathbf{1}\!\left[
\hat{\ell}_i = \ell_i^\star
\right].
\end{equation}

For open-ended task MedBrowseComp~\cite{chen2025medbrowsecomp}, we use rule-based extraction and normalized matching.

\section{Experiments}
\begin{table}[t]
  \centering
  \caption{{Main results on Qwen3-4B-scale backbones under open-ended evaluation.}}
  \label{tab:main-4b-open}
  \resizebox{\textwidth}{!}{%
  \begin{tabular}{@{}lcccccc@{}}
    \toprule
    \textbf{Method} & \textbf{\shortstack{MedBullets\\{\scriptsize op4 / op5}}} & \textbf{\shortstack{MedBrowse\\Comp}} & \textbf{MedXpertQA} & \textbf{\shortstack{MMLU-Pro\\{\scriptsize (health)}}} & \textbf{PubMedQA} & \textbf{Average} \\
    \midrule
    4B Base & 19.8\% / 14.3\% & 0.0\% & 13.3\% & 12.4\% & 17.6\% & 12.9\% \\
    + Agent0~\cite{xia2025agent0} & 25.3\% / 18.8\% & 0.0\% & 10.2\% & 15.2\% & 13.7\% & 13.9\% \\
    + SPIRAL~\cite{liu2025spiral} & 25.6\% / 20.5\% & 4.0\% & 10.2\% & 24.6\% & 24.5\% & 18.2\% \\
    + Absolute Zero~\cite{zhao2025absolutezeroreinforcedselfplay} & 29.3\% / 24.8\% & 0.0\% & 13.2\% & 32.4\% & 27.6\% & 21.2\% \\
    \textbf{+ MedZERO (ours)} & \textbf{41.9\% / 36.0\%} & \textbf{42.0\%} & \textbf{19.8\%} & \textbf{40.2\%} & \textbf{33.5\%} & \textbf{35.6\%} \\
    \bottomrule
  \end{tabular}}
\end{table}

\begin{table}[t]
  \centering
  \caption{{Main results on Qwen3-8B-scale backbones under open-ended evaluation.}}
  \label{tab:main-78b-open}
  \resizebox{\textwidth}{!}{%
  \begin{tabular}{@{}lcccccc@{}}
    \toprule
    \textbf{Method} & \textbf{\shortstack{MedBullets\\{\scriptsize op4 / op5}}} & \textbf{\shortstack{MedBrowse\\Comp}} & \textbf{MedXpertQA} & \textbf{\shortstack{MMLU-Pro\\{\scriptsize (health)}}} & \textbf{PubMedQA} & \textbf{Average} \\
    \midrule
    8B Base & 32.1\% / 23.7\% & 4.0\% & 14.5\% & 57.8\% & 77.5\% & 34.9\% \\
    + Agent0~\cite{xia2025agent0} & 52.4\% / 49.2\% & 4.0\% & 18.0\% & 57.6\% & 81.0\% & 43.7\% \\
    + SPIRAL~\cite{liu2025spiral} & 51.0\% / \textbf{50.2\%} & 4.0\% & 17.2\% & 53.9\% & 77.6\% & 42.3\% \\
    + Absolute Zero~\cite{zhao2025absolutezeroreinforcedselfplay} & 45.1\% / 38.6\% & 0.0\% & 12.5\% & 49.4\% & 76.1\% & 37.0\% \\
    \textbf{+ MedZERO (ours)} & \textbf{55.5\%} / 49.7\% & \textbf{46.0\%} & \textbf{20.8\%} & \textbf{61.2\%} & \textbf{85.2\%} & \textbf{53.6\%} \\
    \bottomrule
  \end{tabular}}
\end{table}

\subsection{Experimental Settings}

\noindent\textbf{Setup.}
Following prior work~\cite{huang2025rzeroselfevolvingreasoningllm}, we adopt Qwen3-4B-Base and Qwen3-8B-Base~\cite{yang2025qwen3} as the base models to study scaling effects within the same model family.
Our framework is built on VeRL~\cite{sheng2025hybridflow}.
For post-optimization curriculum curation, we sample
$K_{\mathrm{cur}}=9$ responses per question and compute
option-level agreement over valid mapped responses.
We retain samples with agreement scores in $[0.3,0.8]$.
Reasoner RL uses $K_{\mathrm{RL}}=4$ rollouts per
training instance.
%
For the Examiner, the grounding reward weight is set to \({\lambda}_{\text{ground}}=0.05\).
For the Reasoner, we use VeRL-Tool~\cite{jiang2025verltool} to support retrieval.

\noindent\textbf{Comparison Methods}
We compare MedZERO against several state-of-the-art self-improvement methods. 1) Base Model: The pre-trained base model without any fine-tuning. 2) Self-Evolving Methods: Absolute Zero~\cite{zhao2025absolutezeroreinforcedselfplay}, SPIRAL~\cite{liu2025spiral}, and Agent0~\cite{xia2025agent0}.

\noindent\textbf{Datasets and Metrics}
MedZERO requires no human-annotated data for training. We evaluate MedZERO on five public medical reasoning benchmarks in an open-ended setting, on two suites of benchmarks: 1) Traditional medical reasoning: PubMedQA~\cite{jin2019pubmedqa}, MMLU-Pro (health)~\cite{wang2024mmlu}; 2) Expert-level medical reasoning: MedBullets~\cite{chen2025benchmarking}, MedBrowseComp~\cite{chen2025medbrowsecomp}, MedXpertQA~\cite{zuo2025medxpertqa}, 
with MedBullets reported under two option settings, op4 and op5.
We report the accuracy (pass@1) across all benchmarks.
Because MedBullets is evaluated under both op4 and op5 settings, the overall average is computed across six reported scores.

\subsection{Results}

Tables~\ref{tab:main-4b-open} and~\ref{tab:main-78b-open} report the main results under the open-ended evaluation protocol. MedZERO achieves the best overall performance at both model scales. At the 4B scale, MedZERO obtains the highest average score of 35.6\%, outperforming Absolute Zero (21.2\%), SPIRAL (18.2\%), Agent0 (13.9\%), and the base model (12.9\%). The same trend holds at the 8B scale, where MedZERO again ranks first with an average score of 53.6\%, ahead of Agent0 (43.7\%), SPIRAL (42.3\%), Absolute Zero (37.0\%), and the base model (34.9\%). Overall, these results show that MedZERO provides the strongest gains over both the base model and prior self-improving baselines for open-ended medical reasoning.

\section{Analysis}

\subsection{Component Analysis}

We study MedZERO along two axes: Reasoner configuration and knowledge persistence policy. We examine whether Reasoner optimization improves performance beyond tool access alone, and whether selective persistence helps beyond storing more knowledge. We highlight two findings.

\begin{figure}[tp]
    \centering
    \includegraphics[width=\textwidth]{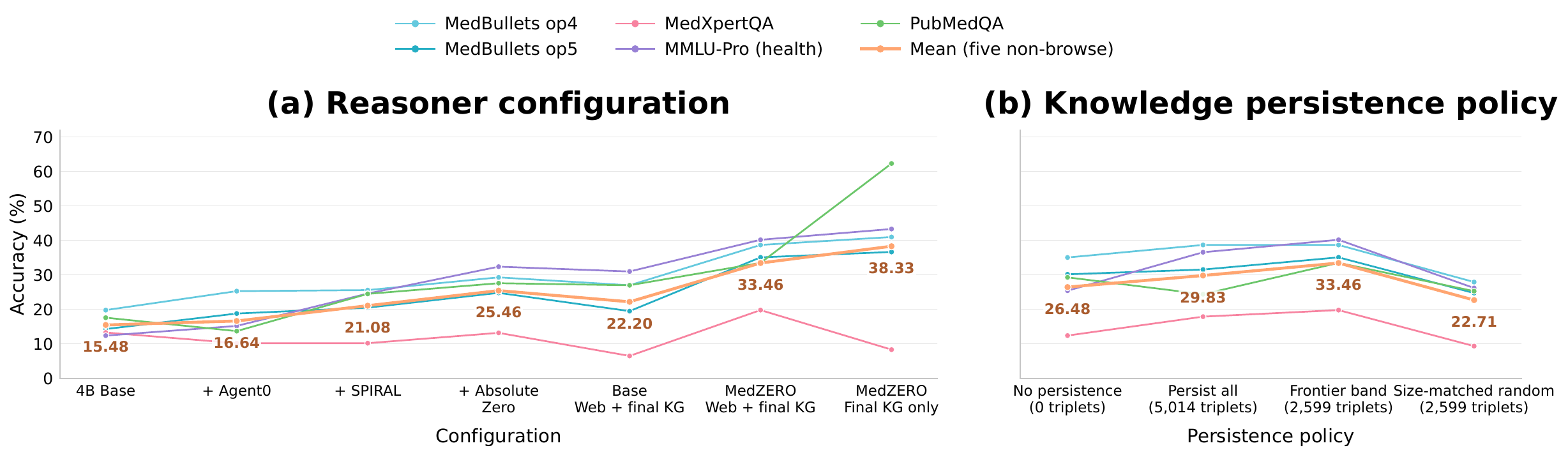}
    \caption{
        Component analysis across (a) Reasoner configurations and
        training baselines, and (b) knowledge persistence policies.
        Scores are reported on MedBullets op4/op5, MedXpertQA,
        MMLU-Pro (health), and PubMedQA.
        The orange curve denotes the unweighted mean of these five
        scores, excluding MedBrowseComp.
        Triplet counts in (b) indicate the size of persistent memory.
    }
    \label{fig:ablation}
\end{figure}

\textbf{(i) Reasoner optimization improves performance beyond tool access alone.}
Figure~\ref{fig:ablation}(a) shows that MedZERO with Web access and the final KG achieves a five-score ex-MedBrowseComp mean of 33.46\%, compared with 22.20\% for the base model with access to the same evidence sources.
It also outperforms the strongest training baseline, Absolute Zero (25.46\%), across all five scores.
These results indicate that MedZERO's gains cannot be explained by tool access alone, supporting the importance of Reasoner optimization.
Removing Web access while retaining the final KG further increases the mean to 38.33\%, but yields uneven changes across benchmarks: PubMedQA improves substantially, whereas MedXpertQA declines.
Thus, the benefit of evidence access is task-dependent rather than uniform.

\textbf{(ii) Selective persistence improves knowledge utility beyond storing more triplets.}
Figure~\ref{fig:ablation}(b) shows that frontier-band admission achieves the strongest results across all five scores, with a mean of 33.46\%, compared with 26.48\% without persistence and 29.83\% when persisting all candidates.
Notably, it retains only 2,599 triplets, versus 5,014 for indiscriminate persistence.
At the same memory size, random selection achieves only 22.71\%, falling below even the no-persistence condition.
These results indicate that the benefit of persistent knowledge depends on which triplets are admitted, rather than on memory size alone, supporting controlled accumulation over indiscriminate storage.

\subsection{Training Stability and Iterative Self-Evolution}


\textbf{Reasoner optimization is stable across training seeds.}
With the Examiner-generated curriculum fixed, we conduct
three independent Reasoner-training runs.
Table~\ref{tab:training-stability}(a) shows that all
benchmark-level sample standard deviations are below
1 percentage point, while the four-score mean is
$36.84 \pm 0.34$\%.
These results support the stability of Reasoner optimization
under a fixed curriculum, rather than measuring variability
across the full self-evolution pipeline.

\textbf{Additional rounds yield further gains.}
Table~\ref{tab:training-stability}(b) shows consistent
improvements over three self-evolution rounds.
MedBullets op4 accuracy increases from 35.20\% to 41.90\%,
while MMLU-Pro improves from 38.94\% to 40.20\%.
These results support the benefit of continued self-evolution
beyond a single round of adaptation.

\begin{table}[t]
    \centering
    \caption{
        Training stability and iterative self-evolution.
        (a) Mean accuracy (\%) $\pm$ sample standard deviation
        across three Reasoner-training seeds with a fixed
        curriculum; standard deviations are in percentage points.
        (b) Accuracy (\%) across three self-evolution rounds,
        averaged over seeds.
    }
    \label{tab:training-stability}

    \begin{minipage}[t]{0.48\textwidth}
        \vspace{0pt}
        \centering
        \textbf{(a) Multi-seed training}
        \par\medskip

        \begin{tabular}{@{}lc@{}}
            \toprule
            Benchmark & Mean $\pm$ Std. \\
            \midrule
            MedBullets op4    & $39.09 \pm 0.68$ \\
            MedBullets op5    & $35.08 \pm 0.98$ \\
            MMLU-Pro (health) & $40.08 \pm 0.32$ \\
            PubMedQA         & $33.10 \pm 0.36$ \\
            \midrule
            Four-score mean  & $36.84 \pm 0.34$ \\
            \bottomrule
        \end{tabular}
    \end{minipage}%
    \hfill
    \begin{minipage}[t]{0.49\textwidth}
        \vspace{0pt}
        \centering
        \textbf{(b) Self-evolution rounds}
        \par\medskip

        \begin{tabular}{@{}lccc@{}}
            \toprule
            Benchmark & R1 & R2 & R3 \\
            \midrule
            MedBullets op4
                & 35.20 & 37.30 & \textbf{41.90} \\
            MMLU-Pro
                & 38.94 & 39.02 & \textbf{40.20} \\
            \bottomrule
        \end{tabular}
    \end{minipage}
\end{table}
\subsection{Evaluation Robustness}

\begin{figure}[t]
    \centering
    \includegraphics[width=\linewidth]
        {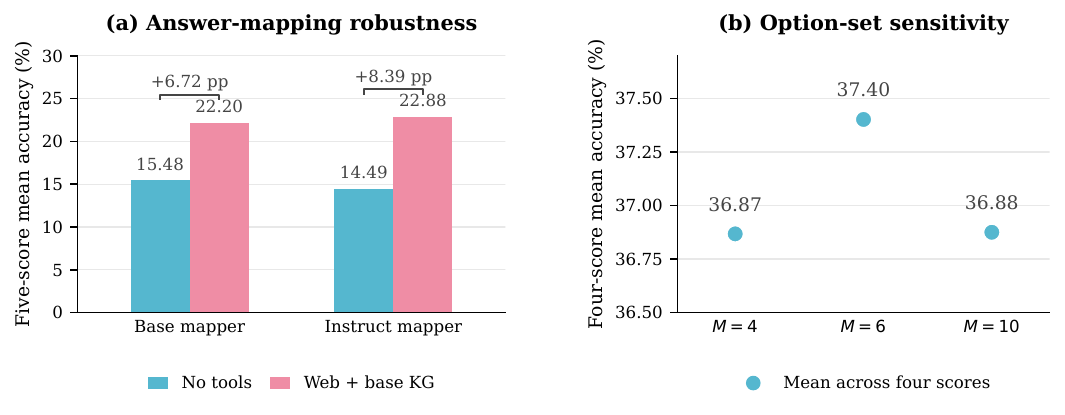}
    \caption{
        Evaluation robustness.
        (a) Remapping identical cached responses with
        Qwen3-4B-Base and Qwen3-4B-Instruct-2507 preserves
        the aggregate gain from Web access and the base KG.
        Brackets indicate gains in percentage points.
        (b) Four-score mean accuracy for option-set cardinalities
$M \in \{4,6,10\}$, each using an independently generated
and filtered curriculum.
    }
    \label{fig:evaluation-robustness}
\end{figure}

\textbf{The audited tool-use gain persists across answer mappers.}
By default, the mapper is implemented with Qwen3-4B-Base. We remap the same cached responses using Qwen3-4B-Instruct-2507.
Figure~\ref{fig:evaluation-robustness}(a) shows that the
gain from Web access and the base KG remains positive
under both mappers: 6.72 and 8.39 percentage points,
respectively.

\textbf{Aggregate performance varies little across option-set sizes.}
The Reasoner receives only the question stem and produces
a free-form response; options are used only for subsequent
verification.
Using independently generated and filtered curricula with
$M \in \{4,6,10\}$ options, the mean scores on MedBullets op4/op5, MMLU-Pro (Health), PubMedQA, ranges
from 36.87\% to 37.40\%
(Figure~\ref{fig:evaluation-robustness}(b)).
This comparison measures end-to-end sensitivity, including
changes in curriculum content and retained size, rather
than isolating option count alone.
Detailed mapping results and option-quality audits are
provided in Appendix~\ref{app:answer-mapping} and
Appendix~\ref{app:option-set}.

\section{Conclusion}

We presented \textbf{MedZERO}, a self-evolving framework for open-ended medical reasoning under partial verifiability. MedZERO co-evolves an \textit{Examiner} for frontier medical challenge generation and a \textit{Reasoner} for evidence-grounded reasoning, while using ephemeral-to-persistent knowledge accumulation to separate exploratory self-generated triplets from curated reusable knowledge. Across five medical reasoning benchmarks and two model scales, MedZERO consistently improves over both base models and self-evolving baselines. These results suggest that explicitly designing curriculum construction, supervision, and knowledge reuse for partial verifiability is important for extending self-evolution beyond easily verifiable domains and toward more realistic medical reasoning settings.

\begin{ack}
The work described in this paper was supported by the
Research Grants Council of the Hong Kong Special
Administrative Region, China, under Project T45-401/22-N.
The authors declare no competing interests.
\end{ack}

\bibliographystyle{plain}
\bibliography{main}
\clearpage

\appendix

\section{Ethics Statement}

This work studies a research framework for open-ended medical reasoning and evaluates it only on academic benchmarks. It is not intended for direct clinical use. Any real-world deployment would require prospective clinical validation across diverse patient populations, regulatory review, continuous monitoring for bias and safety, and mandatory oversight by licensed medical professionals. In particular, errors in retrieval, reasoning, or abstention behavior could cause disproportionate harm in high-stakes settings. Accordingly, \textbf{MedZERO} should not be used for medical diagnosis or treatment decisions, and any future clinical application must comply with applicable medical, legal, and ethical requirements.

\section{Limitations}

MedZERO has several important limitations. First, although the framework is designed for partially verifiable medical reasoning, verification remains incomplete. In many cases, the correctness of a response cannot be established by exact matching alone, and supervision still depends on imperfect signals such as retrieved evidence, consistency checks, and judge-based evaluation. As a result, some reasoning errors may remain undetected during self-evolution.

Second, while MedZERO explicitly separates exploratory knowledge from curated persistent knowledge, this design does not eliminate the risk of error accumulation. Incorrect retrieved evidence, noisy self-generated rationales, or biased curation decisions may still introduce persistent mistakes into future reasoning. The framework reduces this risk but does not fully solve it.

Third, our evaluation is limited to benchmark-based settings and does not establish clinical utility. Academic medical QA benchmarks capture only part of real clinical reasoning, and they do not fully reflect longitudinal patient care, multi-modal evidence, workflow constraints, or the consequences of incorrect recommendations in practice.

Fourth, MedZERO depends on the quality of its external tools and evaluation protocol. Retrieval quality, prompt design, observation budget, and judge behavior can all materially affect outcomes. Although we report controlled settings where possible, some benchmark conditions remain sensitive to implementation choices.

These limitations suggest several directions for future work, including stronger verification mechanisms for open-ended reasoning, more reliable knowledge curation strategies, prospective human evaluation, and broader testing in realistic clinical decision-support settings.

\section{Datasets}

See Table~\ref{tab:eval-datasets} for details.

\begin{table}
  \centering
  \caption{Evaluation datasets used in this work.
  ``HF split'' denotes the split name in the corresponding
  Hugging Face repository, not its role in our experiments.
  Sizes indicate the number of examples used for evaluation.}
  \label{tab:eval-datasets}
  \small
  \begin{tabular}{@{}llrll@{}}
    \toprule
    \textbf{Dataset} & \textbf{HF split} & \textbf{Size}
      & \textbf{Answer format} & \textbf{HF repository} \\
    \midrule
    PubMedQA (PQA labeled)
      & \texttt{train}$^{\dagger}$ & 1{,}000
      & 3-option MCQ & qiaojin/PubMedQA \\
    MedXpertQA (Text)
      & \texttt{test} & 2{,}450
      & 10-option MCQ & TsinghuaC3I/MedXpertQA \\
    MMLU-Pro (Health)
      & \texttt{test} & 818
      & 10-option MCQ & TIGER-Lab/MMLU-Pro \\
    MedBullets (op4)
      & \texttt{op4\_test} & 308
      & 4-option MCQ & mkieffer/Medbullets \\
    MedBullets (op5)
      & \texttt{op5\_test} & 308
      & 5-option MCQ & mkieffer/Medbullets \\
    MedBrowseComp
      & \texttt{test} & 50
      & open-ended & AIM-Harvard/MedBrowseComp \\
    \bottomrule
  \end{tabular}

  \vspace{3pt}
  \begin{minipage}{\textwidth}
    \footnotesize
    $^{\dagger}$\texttt{train} is the official Hugging Face split
    name for PubMedQA's \texttt{pqa\_labeled} configuration;
    all 1{,}000 examples are used exclusively for evaluation.
    MedZERO uses no labeled benchmark data from any of the
    datasets listed above for training.
  \end{minipage}
\end{table}

\section{Experimental details}

\paragraph{Compute.}
We run MedZERO on Linux servers with 4 NVIDIA L20X GPUs. 

\paragraph{Examiner RL (verl).}
Unless otherwise noted, we follow the repository defaults: training with GRPO~\cite{shao2024deepseekmath}, global batch size 8, rollout width 4, max batched tokens 6144, max response length 512, \texttt{trainer.max\_steps}=6, and vLLM memory fraction 0.2 with \texttt{max\_model\_len}=6144. 

\paragraph{Reasoner RL.}
Reasoner training uses GRPO technique with a learning rate $10^{-6}$, FSDP, TP size 1, default train batch size 64, and PPO mini-batch 64, rollout $n=4$, validation batch size 256, maximum number of sequences=256 for rollout.

\paragraph{Base medical KG. }
We use PrimeKG~\citep{chandak2023building} as the base medical knowledge graph.

\section{Systems Efficiency}

We next evaluate whether MedZERO introduces substantial systems overhead.

\begin{figure}[ht]
\centering
\includegraphics[width=0.9\textwidth]{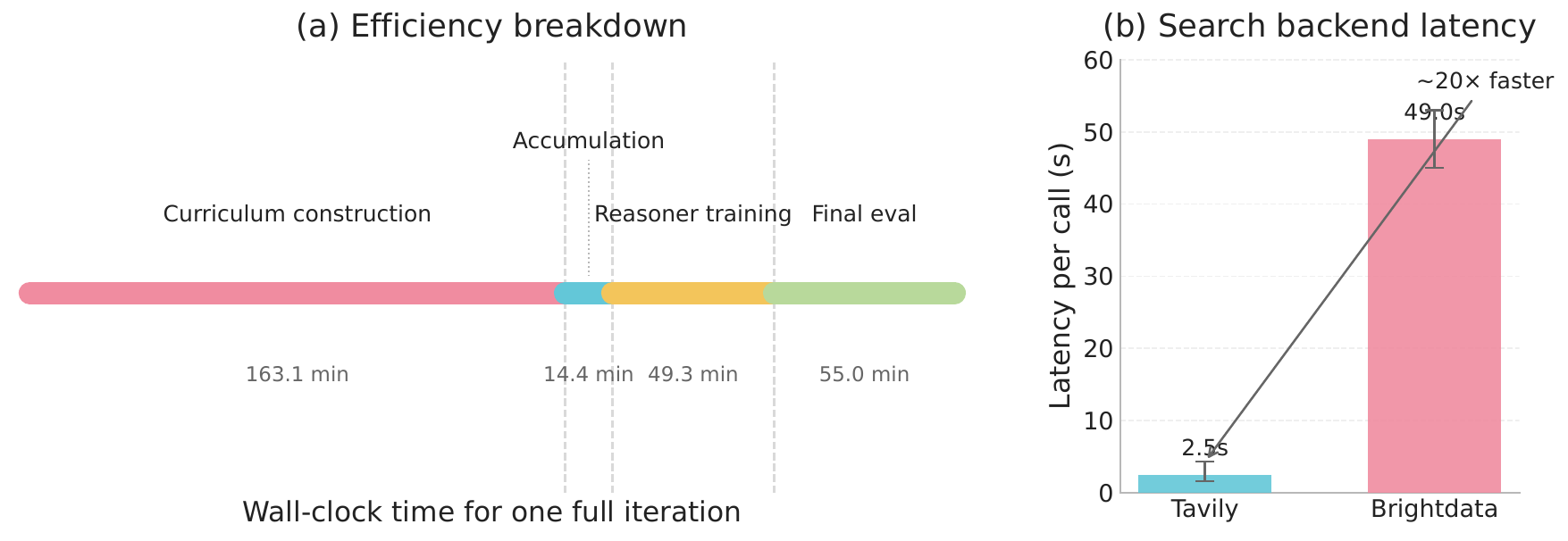}
\caption{(a) Wall-time breakdown for a full MedZERO self-evolution iteration. (b) Web search latency per call (end-to-end wall time, measured in the same execution environment across 5 consecutive calls).}
\label{fig:efficiency}  
\end{figure}

\textbf{(i) Controlled knowledge accumulation is not the main runtime bottleneck.}
Figure~\ref{fig:efficiency}(a) shows that Examiner-side curriculum construction dominates wall-clock time in a full self-evolution iteration, while controlled knowledge accumulation accounts for only a small fraction of the total runtime. Within accumulation, the main cost comes from scoring candidate questions rather than from generation or output materialization.

\textbf{(ii) The evidence-access stack has practical benchmark-scale latency.}
Figure~\ref{fig:efficiency}(b) shows that the FAISS-~\cite{douze2025faiss} and tool-server-based setup of Tavily substantially reduces latency relative to Brightdata. Replacing slower web-search backends with Tavily improves end-to-end latency by roughly an order of magnitude and yields a tighter latency distribution, making evidence-augmented evaluation more predictable at scale.

\section{Answer-Mapping Robustness}
\label{app:answer-mapping}

\paragraph{Open-ended reasoning and post-hoc mapping.}
The Reasoner receives only the question stem and generates
a free-form response. It does not observe the answer options
during response generation.
Answer mapping is performed afterward: a separate evaluation
model receives the generated response and the corresponding
options, and maps the response to an option for scoring.
Thus, the options serve as a downstream verification interface,
rather than an exposed candidate set from which the Reasoner
selects its answer.

\paragraph{Controlled comparison.}
We assess sensitivity to the answer mapper by comparing
Qwen3-4B-Base with Qwen3-4B-Instruct-2507.
For each evaluated system configuration, we reuse exactly
the same cached Reasoner responses under both mappers;
the Reasoner is neither retrained nor queried again.
The corresponding answer options and reference labels
are also unchanged.
This comparison therefore measures sensitivity to the
downstream mapping stage without introducing variation
from response generation.

We evaluate two configurations: a no-tool configuration
and a configuration with Web access and the accumulated knowledge
graph (Web + base KG).
For each mapper, we report the five-score mean accuracy
and the difference between the two configurations.

\begin{table}[t]
    \centering
    \caption{
        Answer-mapping robustness using identical cached
        Reasoner responses for each system configuration.
        Scores are five-score mean accuracy (\%).
        Gain denotes the improvement of Web + base KG over
        the no-tool configuration, in percentage points.
    }
    \label{tab:answer-mapping-detail}
    \begin{tabular}{@{}lccc@{}}
        \toprule
        Mapper & No tools & Web + base KG & Gain (pp) \\
        \midrule
        Base
            & 15.48 & 22.20 & +6.72 \\
        Instruct
            & 14.49 & 22.88 & +8.39 \\
        \bottomrule
    \end{tabular}
\end{table}

\paragraph{Results and interpretation.}
As shown in Table~\ref{tab:answer-mapping-detail},
Web + base KG improves the aggregate score under both
mappers, with gains of 6.72 and 8.39 percentage points,
respectively.
The positive aggregate gain therefore persists across
the two tested mapping models, although its magnitude
changes.

This experiment does not establish that either mapper
is error-free, nor does it demonstrate agreement on
individual responses.
It supports the narrower conclusion that the aggregate
benefit observed in this comparison is not specific to
a single tested answer mapper.

\section{Option-Set Sensitivity}
\label{app:option-set}

\paragraph{Motivation.}
Although the Reasoner does not observe the answer options,
option-set construction may still affect the downstream
verification signal and the examples retained for training.
We therefore evaluate whether varying the number of
Examiner-generated options materially changes the
performance of the resulting Reasoner.

\paragraph{Experimental design.}
We consider option-set cardinalities
$M \in \{4,6,10\}$.
Each setting independently generates and filters its own
complete Examiner curriculum.
We do not construct smaller option sets by removing
distractors from a shared larger set.
The resulting curricula contain 601, 458, and 283 retained
items for $M=4$, $M=6$, and $M=10$, respectively.

Because of computational and time constraints, we evaluate
this sensitivity experiment on four benchmark scores:
MedBullets op4, MedBullets op5, MMLU-Pro (health),
and PubMedQA.
The four-score mean is the unweighted average of these
scores.

\begin{table}[t]
    \centering
    \caption{
        End-to-end sensitivity to option-set cardinality.
        Each setting uses an independently generated and
        filtered curriculum.
        Benchmark scores and the four-score mean are
        accuracy (\%). Averaged over seeds.
    }
    \label{tab:option-set-detail}
    \begin{tabular}{@{}lccc@{}}
        \toprule
        Benchmark / statistic
            & $M=4$ & $M=6$ & $M=10$ \\
        \midrule
        MedBullets op4
            & 38.86 & 38.70 & 38.70 \\
        MedBullets op5
            & 34.06 & 35.10 & 35.10 \\
        MMLU-Pro (health)
            & 41.35 & 42.21 & 40.20 \\
        PubMedQA
            & 33.20 & 33.60 & 33.50 \\
        \midrule
        Four-score mean
            & 36.87 & 37.40 & 36.88 \\
        Retained curriculum items
            & 601 & 458 & 283 \\
        \bottomrule
    \end{tabular}
\end{table}

\paragraph{Performance sensitivity.}
Table~\ref{tab:option-set-detail} shows that the reported
four-score means range from 36.87\% to 37.40\%.
The means for $M=4$ and $M=10$ differ by only
0.01 percentage points.
Increasing the number of options does not consistently
improve performance across benchmarks.
In particular, MMLU-Pro (health) and PubMedQA attain
their highest scores at $M=6$, whereas MedBullets op4
is highest at $M=4$ and MedBullets op5 is tied between
$M=6$ and $M=10$.

\paragraph{Option-quality audit.}
We randomly sample 200 retained examples from each
cardinality setting and audit five properties:
correct-answer coverage, near-equivalent options,
answer leakage, frontier-band membership, and format
validity.
Table~\ref{tab:option-quality-audit} reports the observed
rates within these samples.

\begin{table}[t]
    \centering
    \caption{
        Option-quality audit of 200 randomly sampled
        retained examples per setting.
        All entries are percentages within the audited
        samples, not population-level guarantees.
    }
    \label{tab:option-quality-audit}
    \begin{tabular}{@{}lccc@{}}
        \toprule
        Metric & $M=4$ & $M=6$ & $M=10$ \\
        \midrule
        Correct-answer coverage
            & 100.0 & 100.0 & 100.0 \\
        Near-equivalent-option rate
            & 0.0 & 0.5 & 0.0 \\
        Answer-leakage rate
            & 0.0 & 0.5 & 0.0 \\
        In-band rate
            & 100.0 & 100.0 & 100.0 \\
        Format-valid rate
            & 100.0 & 99.0 & 100.0 \\
        \bottomrule
    \end{tabular}
\end{table}

\paragraph{Audit results.}
All audited samples contain a correct-answer option
and satisfy the frontier-band criterion.
Observed format validity is 99--100\%, while the
near-equivalent-option and answer-leakage rates are
at most 0.5\%.
These findings indicate few observed option-quality
issues among the sampled retained examples.
They do not imply that every retained example is
free of such issues.

\paragraph{Scope and limitations.}
This experiment evaluates end-to-end sensitivity to
option-set cardinality, rather than isolating the causal
effect of option count alone.
Because each setting independently generates and filters
its curriculum, curriculum content and retained size
also vary across settings.
Moreover, the comparison does not provide repeated-run
uncertainty estimates for each cardinality and therefore
does not establish statistical equivalence.
Within these limitations, the small observed variation
in aggregate performance supports robustness across
the three tested option-set settings.

\section{Failure Cases}

\noindent\textbf{Failure case: score does not always align with perceived supervision quality}

\begin{figure}[t]
\centering
\includegraphics[width=1\textwidth]{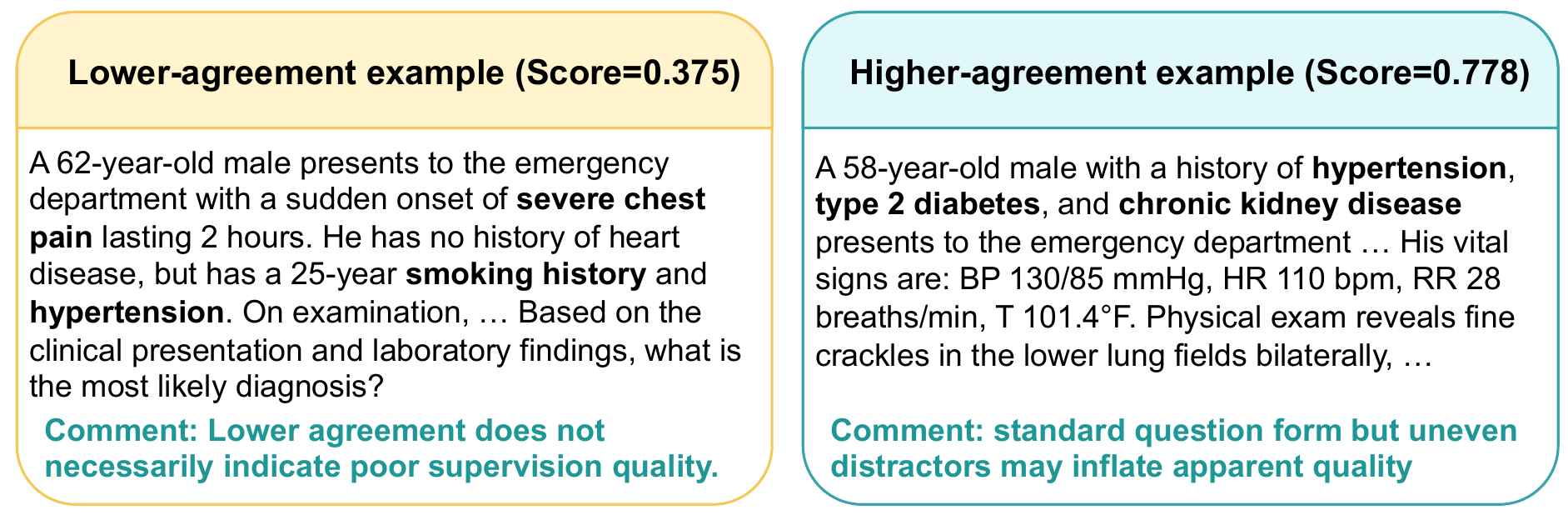}
\caption{Failure cases under controlled knowledge accumulation. These examples illustrate that option-level agreement
is an imperfect proxy for supervision quality.}
\label{fig:case2}
\end{figure}

At the same time, the curation score is not a perfect proxy for human-perceived example quality. Figure~\ref{fig:case2} shows representative cases where score and perceived usefulness do not fully align. Some lower-scoring examples still contain meaningful clinical structure, while some higher-scoring examples may appear cleaner mainly because they better match the reward signal or evaluation heuristic. This failure mode highlights an important limitation of controlled knowledge accumulation: it can reshape the retained supervision distribution in useful ways, but it can also inherit biases or blind spots from the scoring and filtering procedure.

\clearpage
\newpage
\section*{NeurIPS Paper Checklist}

\begin{enumerate}

\item {\bf Claims}
    \item[] Question: Do the main claims made in the abstract and introduction accurately reflect the paper's contributions and scope?
    \item[] Answer: \answerYes{} 
    \item[] Justification: Abstract and Introduction.
    \item[] Guidelines:
    \begin{itemize}
        \item The answer \answerNA{} means that the abstract and introduction do not include the claims made in the paper.
        \item The abstract and/or introduction should clearly state the claims made, including the contributions made in the paper and important assumptions and limitations. A \answerNo{} or \answerNA{} answer to this question will not be perceived well by the reviewers. 
        \item The claims made should match theoretical and experimental results, and reflect how much the results can be expected to generalize to other settings. 
        \item It is fine to include aspirational goals as motivation as long as it is clear that these goals are not attained by the paper. 
    \end{itemize}

\item {\bf Limitations}
    \item[] Question: Does the paper discuss the limitations of the work performed by the authors?
    \item[] Answer: \answerYes{} 
    \item[] Justification: Appendix. B
    \item[] Guidelines:
    \begin{itemize}
        \item The answer \answerNA{} means that the paper has no limitation while the answer \answerNo{} means that the paper has limitations, but those are not discussed in the paper. 
        \item The authors are encouraged to create a separate ``Limitations'' section in their paper.
        \item The paper should point out any strong assumptions and how robust the results are to violations of these assumptions (e.g., independence assumptions, noiseless settings, model well-specification, asymptotic approximations only holding locally). The authors should reflect on how these assumptions might be violated in practice and what the implications would be.
        \item The authors should reflect on the scope of the claims made, e.g., if the approach was only tested on a few datasets or with a few runs. In general, empirical results often depend on implicit assumptions, which should be articulated.
        \item The authors should reflect on the factors that influence the performance of the approach. For example, a facial recognition algorithm may perform poorly when image resolution is low or images are taken in low lighting. Or a speech-to-text system might not be used reliably to provide closed captions for online lectures because it fails to handle technical jargon.
        \item The authors should discuss the computational efficiency of the proposed algorithms and how they scale with dataset size.
        \item If applicable, the authors should discuss possible limitations of their approach to address problems of privacy and fairness.
        \item While the authors might fear that complete honesty about limitations might be used by reviewers as grounds for rejection, a worse outcome might be that reviewers discover limitations that aren't acknowledged in the paper. The authors should use their best judgment and recognize that individual actions in favor of transparency play an important role in developing norms that preserve the integrity of the community. Reviewers will be specifically instructed to not penalize honesty concerning limitations.
    \end{itemize}

\item {\bf Theory assumptions and proofs}
    \item[] Question: For each theoretical result, does the paper provide the full set of assumptions and a complete (and correct) proof?
    \item[] Answer: \answerNA{} 
    \item[] Justification: We do not provide theoretical results in this paper. 
    \item[] Guidelines:
    \begin{itemize}
        \item The answer \answerNA{} means that the paper does not include theoretical results. 
        \item All the theorems, formulas, and proofs in the paper should be numbered and cross-referenced.
        \item All assumptions should be clearly stated or referenced in the statement of any theorems.
        \item The proofs can either appear in the main paper or the supplemental material, but if they appear in the supplemental material, the authors are encouraged to provide a short proof sketch to provide intuition. 
        \item Inversely, any informal proof provided in the core of the paper should be complemented by formal proofs provided in appendix or supplemental material.
        \item Theorems and Lemmas that the proof relies upon should be properly referenced. 
    \end{itemize}

    \item {\bf Experimental result reproducibility}
    \item[] Question: Does the paper fully disclose all the information needed to reproduce the main experimental results of the paper to the extent that it affects the main claims and/or conclusions of the paper (regardless of whether the code and data are provided or not)?
    \item[] Answer: \answerYes{} 
    \item[] Justification: Experiments section 1. 4.1. Experimental Settings 
    \item[] Guidelines:
    \begin{itemize}
        \item The answer \answerNA{} means that the paper does not include experiments.
        \item If the paper includes experiments, a \answerNo{} answer to this question will not be perceived well by the reviewers: Making the paper reproducible is important, regardless of whether the code and data are provided or not.
        \item If the contribution is a dataset and\slash or model, the authors should describe the steps taken to make their results reproducible or verifiable. 
        \item Depending on the contribution, reproducibility can be accomplished in various ways. For example, if the contribution is a novel architecture, describing the architecture fully might suffice, or if the contribution is a specific model and empirical evaluation, it may be necessary to either make it possible for others to replicate the model with the same dataset, or provide access to the model. In general. releasing code and data is often one good way to accomplish this, but reproducibility can also be provided via detailed instructions for how to replicate the results, access to a hosted model (e.g., in the case of a large language model), releasing of a model checkpoint, or other means that are appropriate to the research performed.
        \item While NeurIPS does not require releasing code, the conference does require all submissions to provide some reasonable avenue for reproducibility, which may depend on the nature of the contribution. For example
        \begin{enumerate}
            \item If the contribution is primarily a new algorithm, the paper should make it clear how to reproduce that algorithm.
            \item If the contribution is primarily a new model architecture, the paper should describe the architecture clearly and fully.
            \item If the contribution is a new model (e.g., a large language model), then there should either be a way to access this model for reproducing the results or a way to reproduce the model (e.g., with an open-source dataset or instructions for how to construct the dataset).
            \item We recognize that reproducibility may be tricky in some cases, in which case authors are welcome to describe the particular way they provide for reproducibility. In the case of closed-source models, it may be that access to the model is limited in some way (e.g., to registered users), but it should be possible for other researchers to have some path to reproducing or verifying the results.
        \end{enumerate}
    \end{itemize}

\item {\bf Open access to data and code}
    \item[] Question: Does the paper provide open access to the data and code, with sufficient instructions to faithfully reproduce the main experimental results, as described in supplemental material?
    \item[] Answer: \answerYes{} 
    \item[] Justification: The datasets are publicly available and listed in Appendix C. Code available at \url{https://github.com/IcecreamArtist/MedZERO}. 
    \item[] Guidelines:
    \begin{itemize}
        \item The answer \answerNA{} means that paper does not include experiments requiring code.
        \item Please see the NeurIPS code and data submission guidelines (\url{https://neurips.cc/public/guides/CodeSubmissionPolicy}) for more details.
        \item While we encourage the release of code and data, we understand that this might not be possible, so \answerNo{} is an acceptable answer. Papers cannot be rejected simply for not including code, unless this is central to the contribution (e.g., for a new open-source benchmark).
        \item The instructions should contain the exact command and environment needed to run to reproduce the results. See the NeurIPS code and data submission guidelines (\url{https://neurips.cc/public/guides/CodeSubmissionPolicy}) for more details.
        \item The authors should provide instructions on data access and preparation, including how to access the raw data, preprocessed data, intermediate data, and generated data, etc.
        \item The authors should provide scripts to reproduce all experimental results for the new proposed method and baselines. If only a subset of experiments are reproducible, they should state which ones are omitted from the script and why.
        \item At submission time, to preserve anonymity, the authors should release anonymized versions (if applicable).
        \item Providing as much information as possible in supplemental material (appended to the paper) is recommended, but including URLs to data and code is permitted.
    \end{itemize}

\item {\bf Experimental setting/details}
    \item[] Question: Does the paper specify all the training and test details (e.g., data splits, hyperparameters, how they were chosen, type of optimizer) necessary to understand the results?
    \item[] Answer: \answerYes{} 
    \item[] Justification: Experiment section 1, 4.1. Experimental Settings. 
    \item[] Guidelines:
    \begin{itemize}
        \item The answer \answerNA{} means that the paper does not include experiments.
        \item The experimental setting should be presented in the core of the paper to a level of detail that is necessary to appreciate the results and make sense of them.
        \item The full details can be provided either with the code, in appendix, or as supplemental material.
    \end{itemize}

\item {\bf Experiment statistical significance}
    \item[] Question: Does the paper report error bars suitably and correctly defined or other appropriate information about the statistical significance of the experiments?
    \item[] Answer: \answerYes{} 
    \item[] Justification: Analysis section 5.2. 
    \item[] Guidelines:
    \begin{itemize}
        \item The answer \answerNA{} means that the paper does not include experiments.
        \item The authors should answer \answerYes{} if the results are accompanied by error bars, confidence intervals, or statistical significance tests, at least for the experiments that support the main claims of the paper.
        \item The factors of variability that the error bars are capturing should be clearly stated (for example, train/test split, initialization, random drawing of some parameter, or overall run with given experimental conditions).
        \item The method for calculating the error bars should be explained (closed form formula, call to a library function, bootstrap, etc.)
        \item The assumptions made should be given (e.g., Normally distributed errors).
        \item It should be clear whether the error bar is the standard deviation or the standard error of the mean.
        \item It is OK to report 1-sigma error bars, but one should state it. The authors should preferably report a 2-sigma error bar than state that they have a 96\% CI, if the hypothesis of Normality of errors is not verified.
        \item For asymmetric distributions, the authors should be careful not to show in tables or figures symmetric error bars that would yield results that are out of range (e.g., negative error rates).
        \item If error bars are reported in tables or plots, the authors should explain in the text how they were calculated and reference the corresponding figures or tables in the text.
    \end{itemize}

\item {\bf Experiments compute resources}
    \item[] Question: For each experiment, does the paper provide sufficient information on the computer resources (type of compute workers, memory, time of execution) needed to reproduce the experiments?
    \item[] Answer: \answerYes{} 
    \item[] Justification: Appendix D. Experimental Details. 
    \item[] Guidelines:
    \begin{itemize}
        \item The answer \answerNA{} means that the paper does not include experiments.
        \item The paper should indicate the type of compute workers CPU or GPU, internal cluster, or cloud provider, including relevant memory and storage.
        \item The paper should provide the amount of compute required for each of the individual experimental runs as well as estimate the total compute. 
        \item The paper should disclose whether the full research project required more compute than the experiments reported in the paper (e.g., preliminary or failed experiments that didn't make it into the paper). 
    \end{itemize}
    
\item {\bf Code of ethics}
    \item[] Question: Does the research conducted in the paper conform, in every respect, with the NeurIPS Code of Ethics \url{https://neurips.cc/public/EthicsGuidelines}?
    \item[] Answer: \answerYes{} 
    \item[] Justification: Yes.
    \item[] Guidelines:
    \begin{itemize}
        \item The answer \answerNA{} means that the authors have not reviewed the NeurIPS Code of Ethics.
        \item If the authors answer \answerNo, they should explain the special circumstances that require a deviation from the Code of Ethics.
        \item The authors should make sure to preserve anonymity (e.g., if there is a special consideration due to laws or regulations in their jurisdiction).
    \end{itemize}

\item {\bf Broader impacts}
    \item[] Question: Does the paper discuss both potential positive societal impacts and negative societal impacts of the work performed?
    \item[] Answer: \answerYes{} 
    \item[] Justification: Appendix A. Ethics Statement 
    \item[] Guidelines:
    \begin{itemize}
        \item The answer \answerNA{} means that there is no societal impact of the work performed.
        \item If the authors answer \answerNA{} or \answerNo, they should explain why their work has no societal impact or why the paper does not address societal impact.
        \item Examples of negative societal impacts include potential malicious or unintended uses (e.g., disinformation, generating fake profiles, surveillance), fairness considerations (e.g., deployment of technologies that could make decisions that unfairly impact specific groups), privacy considerations, and security considerations.
        \item The conference expects that many papers will be foundational research and not tied to particular applications, let alone deployments. However, if there is a direct path to any negative applications, the authors should point it out. For example, it is legitimate to point out that an improvement in the quality of generative models could be used to generate Deepfakes for disinformation. On the other hand, it is not needed to point out that a generic algorithm for optimizing neural networks could enable people to train models that generate Deepfakes faster.
        \item The authors should consider possible harms that could arise when the technology is being used as intended and functioning correctly, harms that could arise when the technology is being used as intended but gives incorrect results, and harms following from (intentional or unintentional) misuse of the technology.
        \item If there are negative societal impacts, the authors could also discuss possible mitigation strategies (e.g., gated release of models, providing defenses in addition to attacks, mechanisms for monitoring misuse, mechanisms to monitor how a system learns from feedback over time, improving the efficiency and accessibility of ML).
    \end{itemize}
    
\item {\bf Safeguards}
    \item[] Question: Does the paper describe safeguards that have been put in place for responsible release of data or models that have a high risk for misuse (e.g., pre-trained language models, image generators, or scraped datasets)?
    \item[] Answer: \answerYes{} 
    \item[] Justification: We stated this in the Ethics Statement section. 
    \item[] Guidelines:
    \begin{itemize}
        \item The answer \answerNA{} means that the paper poses no such risks.
        \item Released models that have a high risk for misuse or dual-use should be released with necessary safeguards to allow for controlled use of the model, for example by requiring that users adhere to usage guidelines or restrictions to access the model or implementing safety filters. 
        \item Datasets that have been scraped from the Internet could pose safety risks. The authors should describe how they avoided releasing unsafe images.
        \item We recognize that providing effective safeguards is challenging, and many papers do not require this, but we encourage authors to take this into account and make a best faith effort.
    \end{itemize}

\item {\bf Licenses for existing assets}
    \item[] Question: Are the creators or original owners of assets (e.g., code, data, models), used in the paper, properly credited and are the license and terms of use explicitly mentioned and properly respected?
    \item[] Answer: \answerYes{} 
    \item[] Justification: We have cite all related used datasets in this work. 
    \item[] Guidelines:
    \begin{itemize}
        \item The answer \answerNA{} means that the paper does not use existing assets.
        \item The authors should cite the original paper that produced the code package or dataset.
        \item The authors should state which version of the asset is used and, if possible, include a URL.
        \item The name of the license (e.g., CC-BY 4.0) should be included for each asset.
        \item For scraped data from a particular source (e.g., website), the copyright and terms of service of that source should be provided.
        \item If assets are released, the license, copyright information, and terms of use in the package should be provided. For popular datasets, \url{paperswithcode.com/datasets} has curated licenses for some datasets. Their licensing guide can help determine the license of a dataset.
        \item For existing datasets that are re-packaged, both the original license and the license of the derived asset (if it has changed) should be provided.
        \item If this information is not available online, the authors are encouraged to reach out to the asset's creators.
    \end{itemize}

\item {\bf New assets}
    \item[] Question: Are new assets introduced in the paper well documented and is the documentation provided alongside the assets?
    \item[] Answer: \answerNA{} 
    \item[] Justification: We do not release new assets. 
    \item[] Guidelines:
    \begin{itemize}
        \item The answer \answerNA{} means that the paper does not release new assets.
        \item Researchers should communicate the details of the dataset\slash code\slash model as part of their submissions via structured templates. This includes details about training, license, limitations, etc. 
        \item The paper should discuss whether and how consent was obtained from people whose asset is used.
        \item At submission time, remember to anonymize your assets (if applicable). You can either create an anonymized URL or include an anonymized zip file.
    \end{itemize}

\item {\bf Crowdsourcing and research with human subjects}
    \item[] Question: For crowdsourcing experiments and research with human subjects, does the paper include the full text of instructions given to participants and screenshots, if applicable, as well as details about compensation (if any)? 
    \item[] Answer: \answerNA{} 
    \item[] Justification: The paper does not involve crowdsourcing nor research with human subjects.
    \item[] Guidelines:
    \begin{itemize}
        \item The answer \answerNA{} means that the paper does not involve crowdsourcing nor research with human subjects.
        \item Including this information in the supplemental material is fine, but if the main contribution of the paper involves human subjects, then as much detail as possible should be included in the main paper. 
        \item According to the NeurIPS Code of Ethics, workers involved in data collection, curation, or other labor should be paid at least the minimum wage in the country of the data collector. 
    \end{itemize}

\item {\bf Institutional review board (IRB) approvals or equivalent for research with human subjects}
    \item[] Question: Does the paper describe potential risks incurred by study participants, whether such risks were disclosed to the subjects, and whether Institutional Review Board (IRB) approvals (or an equivalent approval/review based on the requirements of your country or institution) were obtained?
    \item[] Answer: \answerNA{} 
    \item[] Justification: The paper does not involve crowdsourcing nor research with human subjects.
    \item[] Guidelines:
    \begin{itemize}
        \item The answer \answerNA{} means that the paper does not involve crowdsourcing nor research with human subjects.
        \item Depending on the country in which research is conducted, IRB approval (or equivalent) may be required for any human subjects research. If you obtained IRB approval, you should clearly state this in the paper. 
        \item We recognize that the procedures for this may vary significantly between institutions and locations, and we expect authors to adhere to the NeurIPS Code of Ethics and the guidelines for their institution. 
        \item For initial submissions, do not include any information that would break anonymity (if applicable), such as the institution conducting the review.
    \end{itemize}

\item {\bf Declaration of LLM usage}
    \item[] Question: Does the paper describe the usage of LLMs if it is an important, original, or non-standard component of the core methods in this research? Note that if the LLM is used only for writing, editing, or formatting purposes and does \emph{not} impact the core methodology, scientific rigor, or originality of the research, declaration is not required.
    \item[] Answer: \answerYes{} 
    \item[] Justification: 4.1. Experimental Settings. 
    \item[] Guidelines:
    \begin{itemize}
        \item The answer \answerNA{} means that the core method development in this research does not involve LLMs as any important, original, or non-standard components.
        \item Please refer to our LLM policy in the NeurIPS handbook for what should or should not be described.
    \end{itemize}

\end{enumerate}

\end{document}